\documentclass{article} 
\usepackage{iclr2026_conference,times}

\usepackage{amsmath,amsfonts,bm}

\def\eqref#1{equation~\ref{#1}}

\def\1{\bm{1}}

\DeclareMathAlphabet{\mathsfit}{\encodingdefault}{\sfdefault}{m}{sl}
\SetMathAlphabet{\mathsfit}{bold}{\encodingdefault}{\sfdefault}{bx}{n}

\usepackage{hyperref}
\usepackage{url}
\usepackage{graphicx}
\usepackage{amsmath,amssymb,amsfonts}
\usepackage{algorithm}
\usepackage{algorithmic}
\usepackage{booktabs,cleveref}
\usepackage{xcolor} 
\usepackage{wrapfig} 
\usepackage{enumitem} 
\usepackage{multirow} 

\newif\ifcamera
\newif\ifanonymous
\anonymousfalse \cameratrue

\title{Natural Building Blocks for Structured World Models: Theory, Evidence, and Scaling}

\author{
\begin{minipage}{\textwidth}\centering
\quad Lancelot Da Costa\textsuperscript{1,3} \quad\quad\quad
Sanjeev Namjoshi\textsuperscript{1}\\
Mohammed Abbas Ansari\textsuperscript{2}\quad\quad
Bernhard Schölkopf\textsuperscript{3,4} 
\end{minipage}\\
\begin{minipage}{\textwidth}\centering
\vspace{4pt}
\textsuperscript{1}VERSES AI Research Lab \quad
\textsuperscript{2}University of Tübingen\\
\textsuperscript{3}ELLIS Institute, Tübingen \quad
\textsuperscript{4}MPI for Intelligent Systems, Tübingen
\end{minipage}
}

\ifcamera
    \iclrfinalcopy 
\fi

\begin{document}

\maketitle

\thispagestyle{plain}
\pagestyle{plain}              

\vspace{-12pt}
\begin{abstract}
The field of world modeling is fragmented, with researchers developing bespoke architectures that rarely build upon each other.
We propose a framework that specifies the natural building blocks for structured world models based on the fundamental stochastic processes that any world model must capture: discrete processes (logic, symbols) and continuous processes (physics, dynamics); the world model is then defined by the hierarchical composition of these building blocks.
We examine Hidden Markov Models (HMMs) and switching linear dynamical systems (sLDS) as natural building blocks for discrete and continuous modeling---which become partially-observable Markov decision processes (POMDPs) and controlled sLDS when augmented with actions.
This modular approach supports both passive modeling (generation, forecasting) and active control (planning, decision-making) within the same architecture.
We avoid the combinatorial explosion of traditional structure learning by largely fixing the causal architecture and searching over only four depth parameters.
We review practical expressiveness through multimodal generative modeling (passive) and planning from pixels (active), with performance competitive to neural approaches while maintaining interpretability.
The core outstanding challenge is scalable joint structure-parameter learning; current methods finesse this by cleverly growing structure and parameters incrementally, but are limited in their scalability.
If solved, these natural building blocks could provide foundational infrastructure for world modeling, analogous to how standardized layers enabled progress in deep learning.
\end{abstract}

\begin{figure}[h!]
    \centering
    \includegraphics[width=0.91\columnwidth]{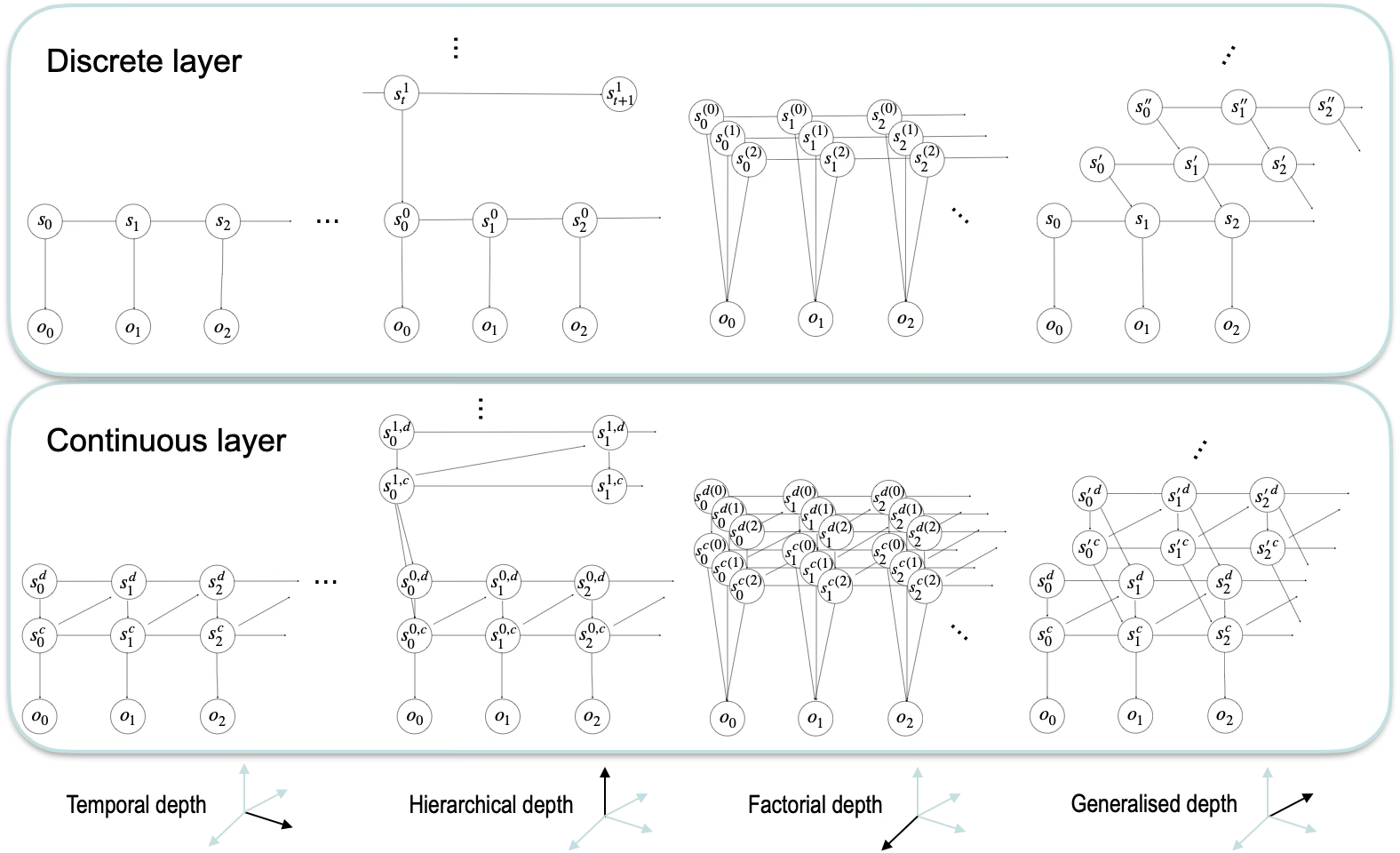}
    \caption{\textbf{Discrete and continuous building blocks demonstrating four types of structural depth.} HMM (discrete) and recurrent sLDS (continuous) layers.
    Temporal depth unfolds dynamics over time, hierarchical depth creates abstraction levels, factorial depth separates independent variation sources, and generalized depth encodes latent memory through higher-order dynamics.}
    \label{fig:architecture}
\end{figure}

\vspace{-0.45cm}
\section{Introduction}
\vspace{-0.25cm}

World modeling \citep{world_modeling_review} suffers from fragmentation---each research group builds custom architectures with little code reuse, mirroring deep learning before frameworks such as Caffe and Keras standardized layers and composition rules \citep{chollet2015keras}.
While neural world models \citep{dreamer, planet} achieve strong performance, they operate as black boxes; on the other hand, structured models offer interpretability but traditionally lack expressiveness (related work in Appendix~\ref{app:related}).

We propose that world models should be built from the fundamental stochastic processes they must capture. Most phenomena in the natural
and complex systems sciences are modeled through
Markov chains, stochastic differential
equations (SDEs) (particularly diffusion processes), and their hierarchical compositions (multi-scale phenomena). We propose that the core building blocks for world modeling should be learnable approximations to these processes. Here, we focus on natural candidates: HMMs \citep{rabiner_hmm_intro} and sLDS \citep{switching_lds_1}, which transform into POMDPs and controlled sLDS when augmented with actions.

Our contributions are: (1) proposing world
  models be built from learnable approximations
   of the fundamental stochastic processes encountered in nature, (2)
  natural building blocks unifying passive
  modeling and active control, (3) four types
  of structural depth finessing combinatorial
  explosion of structure learning while retaining expressiveness, (4)
  reviewing empirical evidence: multimodal
  generation without neural networks,
  behavioral/neural modeling,
  planning from pixels competitive with state-of-the-art, and (5) identifying the
   core challenge for this technology to become foundational infrastructure for world modeling: scalable joint structure-parameter learning.

\vspace{-0.35cm}
\section{Natural Building Blocks for World Models}
\vspace{-0.25cm}

\subsection{Building Blocks}
\vspace{-0.15cm}
We propose two complementary building blocks and describe their hierarchical compositions to capture the essential stochastic processes in world modeling:

\noindent\textbf{1. Hidden Markov Models (HMMs).} HMMs capture discrete partially-observed dynamics, providing a fundamental model for time series (see Appendix~\ref{app:math}). When augmented with control states via generalized depth (Section~\ref{sec:depth}), they transform into POMDPs for agent modeling. These naturally handle logical reasoning, symbolic manipulation, and categorical decisions. While extremely expressive through quantization, they scale poorly to high dimensions.

\noindent\textbf{2. Switching linear dynamical systems (sLDS).} sLDS provide a continuous alternative to HMMs, approximating nonlinear dynamics through switching linear systems (see Appendix~\ref{app:math}). Adding control states through generalized depth enables continuous control. A recent key innovation is the \emph{recurrent} connection where continuous states influence discrete switching (i.e., rsLDS; \citet{linderman_bayesian_2017})---important for contact physics, such as bouncing. While less universal than HMMs (approximating SDEs), their continuous state-spaces scale naturally to higher-dimensional inputs.

\noindent\textbf{Hierarchical Composition.} Higher-level states provide (stochastic) initial conditions for lower-level trajectories, enabling multi-scale modeling; from abstract planning to detailed execution. The framework supports discrete→discrete (logical hierarchies), discrete→continuous (plans to physics), and continuous→continuous (multi-scale physics) compositions.

\vspace{-0.2cm}
\subsection{Four Types of Structural Depth}
\label{sec:depth}
\vspace{-0.15cm}
These building blocks support four complementary types of structural depth (Figure~\ref{fig:architecture}):

\begin{enumerate}[leftmargin=*,topsep=0pt,itemsep=0pt]
\item \textbf{Temporal depth}: Standard time unrolling $(t = 1, 2, 3, ...)$ for sequential dynamics.
\item \textbf{Hierarchical depth}: Multiple abstraction levels, from high-level goals/concepts to detailed ones.
\item \textbf{Factorial depth}: Independent variation sources (e.g., object position, color, texture) that evolve separately.
\item \textbf{Generalized depth}: Memory by extending the latent space to incorporate higher-order latent variables; that is, velocity, acceleration, and higher-order motion, and their discrete counterparts. This extends classical HMMs and rsLDS to express non-Markovian (technically semi-Markovian) dynamics in latent space. These generalized latent states can optionally be defined as controllable enabling action: for example, controlling the first generalized order in an HMM makes it a POMDP, and in an rsLDS makes it a controlled rsLDS.
\end{enumerate}

\vspace{-0.15cm}
\subsection{Inference, Learning \& Structure learning}
\vspace{-0.15cm}
The building blocks support efficient Bayesian inference through variational methods, providing uncertainty quantification over states and parameters \citep{winn_variational_2005,da_costa_active_2020,linderman_bayesian_2017}. However, de novo structure \& Bayesian parameter learning in deep models remains a core challenge.
Fast Structure Learning (FSL; \cite{friston_pixels_2024}) tackles this challenge by incrementally growing structure and parameters jointly, however it is currently limited in scale.


\vspace{-0.35cm}
\section{Theoretical Features}
\vspace{-0.25cm}
\noindent\textbf{Expressiveness.}
HMMs with generalized depth express partially-observable, finite-memory (i.e., semi-Markovian) discrete processes and approximate their continuous counterparts through quantization. rsLDS approximate partially-observable nonlinear diffusion processes through recurrent switching linear dynamics. With generalized depth, rsLDS could approximate non-Markovian SDEs, e.g., \citep{da_costa_theory_2025}, but this implementation remains future work (Section \ref{sec: current state}). Hierarchical composition extends this expressiveness to multi-scale phenomena (analysis in Appendix~\ref{app:math}).

\begin{wrapfigure}{r}{0.45\textwidth}
  \centering
  \vspace{-0.6cm}
  \includegraphics[width=0.43\textwidth]{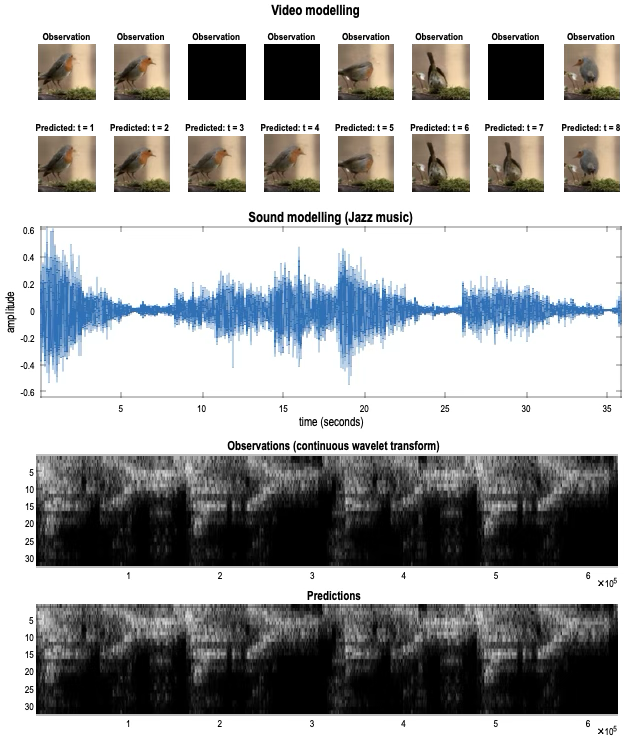}
  \caption{\textbf{Multimodal generative modeling without neural networks.}
  Video (top) and jazz audio generation (bottom) using hierarchical generalized HMMs with FSL. Adapted from \citep{friston_pixels_2024}.}
  \vspace{-1cm}
  \label{fig:multimodal}
\end{wrapfigure}

\noindent\textbf{Avoiding Combinatorial Explosion.}
Traditional structure learning \citep{koller_friedman} faces super-exponential growth ($>10^{12}$ unlabeled DAGs for just 10 variables).
We finesse this by largely fixing the model structure (hierarchical HMMs/rsLDS) and searching only over four depth parameters (Figure \ref{fig:architecture}), reducing the search space dramatically while maintaining expressiveness.

\vspace{-0.35cm}
\section{Empirical Evidence}
\label{sec:evidence}
\vspace{-0.25cm}

\noindent\textbf{Multimodal Generation (Passive).}
Figure~\ref{fig:multimodal} shows coherent video and jazz audio generation without neural networks using hierarchical generalized HMMs and FSL \citep{friston_pixels_2024}, demonstrating multimodal expressiveness with interpretable components. 

\noindent\textbf{Planning from Pixels (Active).} \cite{heins_axiom_2025} demonstrates planning in hierarchical sLDS with model structure learned online with FSL in Atari-style environments, achieving performance \emph{competitive with state-of-the-art neural approaches}: DreamerV3 \citep{dreamerv3} and BBF \citep{bbf}, with better early learning curves, significantly fewer parameters, lower per-step model-update time, and lower end-to-end (including planning) per-step runtime than DreamerV3.

\begin{figure}[t]
  \vspace{-0.3cm}
  \centering
  \includegraphics[width=0.9\columnwidth]{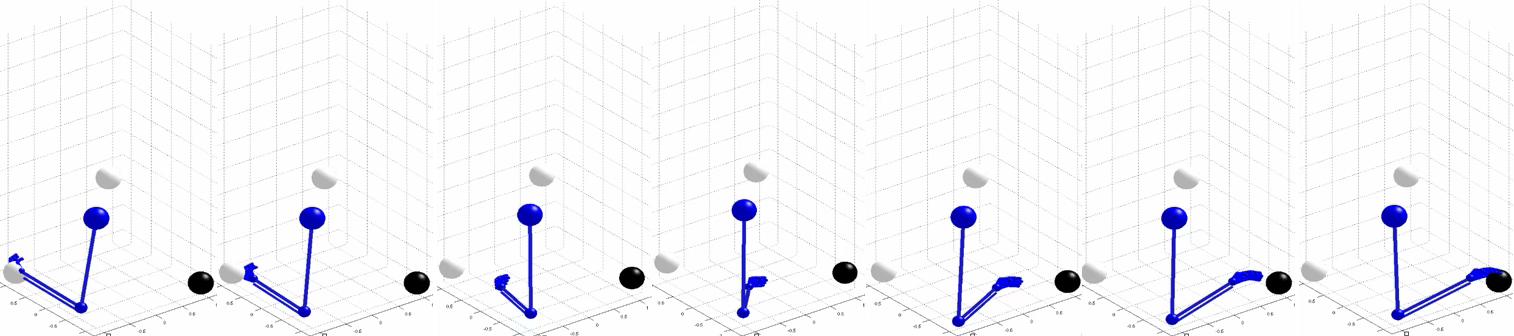}
  \caption{\textbf{Behavioral modeling with discrete→continuous hierarchies.} Motor movement with a hierarchy of two HMM layers followed by a continuous SDE layer with generalized depth. Studies across robotics and cognitive science show a wide range of other behaviors simulated in our proposed model space (Section \ref{sec:evidence}). Figure adapted from \citep{parr_computational_2021}.}
  \label{fig:behavioral}
\end{figure}

\noindent\textbf{Robotics and Behavioral Modeling (Active).}\label{sub:roboticsBehaviour} \citet{parr_computational_2021} demonstrates coherent motor control with discrete→continuous hierarchies (Figure~\ref{fig:behavioral}). \citep{lanillos_active_2021,da_costa_active_2022,da_costa_active_2020} review behavioral studies across robotics and cognitive science where a wide range of behaviors are simulated using models in our proposed space: e.g., from robot navigation \citep{catal_robot_2021}, to robot control \citep{pezzato_mobile_2025}, to eye movements \citep{parr_active_2018}. In all of these works, however, the generative model structure was hard-coded, not learned.

\noindent\textbf{Neural Modeling (Active).}
Proposed model space and inference computations align with recent theoretical and empirical evidence that biological neurons implement active inference on POMDP generative models \citep{isomura}. Extending this formal and empirical link to structure learning holds promise for devising better learning algorithms.

\vspace{-0.35cm}
\section{Current State and Challenges}
\vspace{-0.25cm}
\label{sec: current state}

\begin{table}[h]
\vspace{-0.3cm}
\centering
\caption{Current capabilities and future directions for natural world models.}
\label{tab:capabilities}
\small
\begin{tabular}{cll}
\toprule
\textbf{Status} & \textbf{Capability} & \textbf{Notes} \\
\midrule

\multirow{6}{*}{\checkmark} 
& Interpretable (sparsity, hierarchy) & Structured generative models and FSL \citep{friston_pixels_2024} \\
& Hybrid discrete-continuous modeling & Behavioral modeling \citep{parr_computational_2021,da_costa_active_2020}\\
& Multimodal without neural networks & Generative modeling of video and music \citep{friston_pixels_2024} \\
& Continual learning & Parameter and structure learning via Bayes \citep{heins_axiom_2025}\\
& Sample efficient learning & Via information gain maximizing action selection \citep{heins_axiom_2025} \\
& Parameter efficiency & RL using hierarchical sLDS and FSL \citep{heins_axiom_2025} \\

\midrule

\multirow{3}{*}{Partial}
& Scaling & Need higher dimensional environments \\
& Structure learning & Need more scalable structure learning algorithms\\
& End-to-end uncertainty & Have for states \& parameters but not structure \citep{da_costa_possible_2024}\\

\midrule

\multirow{5}{*}{Future}
& Safety demonstrations & Leveraging uncertainty \citep{da_costa_how_2022,bengio_superintelligent_2025} \\
& Transfer learning \& OOD generalization & Leveraging sparse representations \\
& Theoretical guarantees & Rigorous approximation theorems for building blocks \\
& Open-ended learning & Relies on developing universally expressive model space \\
& Scientific discovery & Active learning on expressive model space \citep{jain_gflownets_2023} \\

\bottomrule
\end{tabular}
\vspace{-0.3cm}
\end{table}

\noindent\textbf{Scalable joint structure \& parameter learning.}
This remains the core challenge, with
  many possible approaches, e.g. \citep{deleu2023joint}. 
  One promising
  avenue casts structure and parameter
  inference as hierarchical Bayesian inference, leveraging uncertainty in action selection to actively learn these with minimal
  samples \citep{murphy_active_nodate,da_costa_possible_2024}. However, this requires
  solving deep Bayesian parameter learning in
  these hierarchies—currently an outstanding
  challenge.

\noindent\textbf{Refining space of models.} We need to identify which variant of sLDS (e.g., \cite{switching_lds_1,linderman_bayesian_2017,heins_axiom_2025}) best balances expressiveness with inference and learnability at scale. Extending sLDS with generalized coordinates
\cite{da_costa_theory_2025} would enable 
modeling of non-Markovian SDEs common in 
physics and complex systems. Finally, we need theoretical assessments of our model space's
   expressiveness (e.g., relative to Turing completeness), and empirical studies to determine whether any expressiveness gap limits practical applications.

\noindent\textbf{AI for Complex Systems Modeling.} Since our building blocks approximate the most common processes encountered in the natural and complex systems sciences, we envision that scalable structure learning algorithms could automatically recover many of these scientific models—enabling AI-driven scientific discovery \citep{jain_gflownets_2023}.

\vspace{-0.4cm}
\section{Conclusion}
\vspace{-0.25cm}

We proposed that world models be built
   compositionally from the fundamental
  stochastic processes they must capture:
  discrete processes for logic and symbols, and
   continuous processes for physics and
  dynamics. The most fundamental processes in
  these classes—Markov chains and stochastic differential
  equations—underlie much of modeling in the natural and
  complex systems sciences. We therefore
  investigated natural, learnable
  approximations to these fundamental
  processes: generalized HMMs and sLDS as
  building blocks.

  This approach enables theoretical and
  practical expressiveness while avoiding the
  combinatorial explosion of traditional
  structure learning. Our review of empirical evidence
  demonstrates competitive performance in planning from pixels with neural world models while maintaining
  interpretability.

  The core challenge remains scalable joint
  structure-parameter learning. If solved,
  these building blocks could provide
  foundational infrastructure for world
  modeling—analogous to how standardized layers
   enabled progress in deep learning. The
  combination of interpretability, uncertainty
  quantification, and expressiveness would
  unlock applications in safety-critical
  domains and scientific discovery. We invite
  the community to join us in developing
  scalable learning algorithms on these foundations, bridging the gap
  between symbolic and neural
  world modeling.

\vspace{-0.3cm}

\ifcamera
    \section*{Acknowledgments}
    The authors thank Alexander Tschantz, Christopher L. Buckley and the VERSES Machine Learning Foundations group for helpful discussions.
\fi

\bibliographystyle{iclr2026_conference}
\bibliography{wmw_paper,references/bib}

\begin{thebibliography}{51}
\providecommand{\natexlab}[1]{#1}
\providecommand{\url}[1]{\texttt{#1}}
\expandafter\ifx\csname urlstyle\endcsname\relax
  \providecommand{\doi}[1]{doi: #1}\else
  \providecommand{\doi}{doi: \begingroup \urlstyle{rm}\Url}\fi

\bibitem[Bengio et~al.(2025)Bengio, Cohen, Fornasiere, Ghosn, Greiner, MacDermott, Mindermann, Oberman, Richardson, Richardson, Rondeau, St-Charles, and Williams-King]{bengio_superintelligent_2025}
Yoshua Bengio, Michael Cohen, Damiano Fornasiere, Joumana Ghosn, Pietro Greiner, Matt MacDermott, Sören Mindermann, Adam Oberman, Jesse Richardson, Oliver Richardson, Marc-Antoine Rondeau, Pierre-Luc St-Charles, and David Williams-King.
\newblock Superintelligent {Agents} {Pose} {Catastrophic} {Risks}: {Can} {Scientist} {AI} {Offer} a {Safer} {Path}?, February 2025.
\newblock URL \url{http://arxiv.org/abs/2502.15657}.
\newblock arXiv:2502.15657 [cs].

\bibitem[Bishop(1994)]{mixture_density}
Christopher~M. Bishop.
\newblock Mixture density networks.
\newblock Technical report, Aston University Neural Computing Research Group, Birmingham, United Kingdom, 1994.
\newblock Available online at \url{https://publications.aston.ac.uk/id/eprint/373/1/NCRG_94_004.pdf}.

\bibitem[Castelletti et~al.(2024)Castelletti, Consonni, and Della~Vedova]{joint_structure_params}
Federico Castelletti, Guido Consonni, and Marco~L Della~Vedova.
\newblock Joint structure learning and causal effect estimation for categorical graphical models.
\newblock \emph{Biometrics}, 80\penalty0 (3):\penalty0 ujae067, 07 2024.
\newblock ISSN 0006-341X.
\newblock \doi{10.1093/biomtc/ujae067}.
\newblock URL \url{https://doi.org/10.1093/biomtc/ujae067}.

\bibitem[Chang \& Athans(1978)Chang and Athans]{switching_lds_2}
C.~B. Chang and M.~Athans.
\newblock State estimation for discrete systems with switching parameters.
\newblock \emph{IEEE Transactions on Aerospace and Electronic Systems}, AES-14\penalty0 (3):\penalty0 418--425, 1978.
\newblock \doi{10.1109/TAES.1978.308603}.

\bibitem[Chollet et~al.(2015)]{chollet2015keras}
Fran\c{c}ois Chollet et~al.
\newblock Keras.
\newblock \url{https://keras.io}, 2015.

\bibitem[Da~Costa et~al.(2020)Da~Costa, Parr, Sajid, Veselic, Neacsu, and Friston]{da_costa_active_2020}
Lancelot Da~Costa, Thomas Parr, Noor Sajid, Sebastijan Veselic, Victorita Neacsu, and Karl Friston.
\newblock Active inference on discrete state-spaces: {A} synthesis.
\newblock \emph{Journal of Mathematical Psychology}, 99:\penalty0 102447, December 2020.
\newblock ISSN 0022-2496.
\newblock \doi{10.1016/j.jmp.2020.102447}.
\newblock URL \url{http://www.sciencedirect.com/science/article/pii/S0022249620300857}.

\bibitem[Da~Costa et~al.(2022{\natexlab{a}})Da~Costa, Lanillos, Sajid, Friston, and Khan]{da_costa_how_2022}
Lancelot Da~Costa, Pablo Lanillos, Noor Sajid, Karl Friston, and Shujhat Khan.
\newblock How {Active} {Inference} {Could} {Help} {Revolutionise} {Robotics}.
\newblock \emph{Entropy}, 24\penalty0 (3):\penalty0 361, March 2022{\natexlab{a}}.
\newblock ISSN 1099-4300.
\newblock \doi{10.3390/e24030361}.
\newblock URL \url{https://www.mdpi.com/1099-4300/24/3/361}.
\newblock tex.ids= dacostaHowActiveInference2022 number: 3 publisher: Multidisciplinary Digital Publishing Institute.

\bibitem[Da~Costa et~al.(2022{\natexlab{b}})Da~Costa, Tenka, Zhao, and Sajid]{da_costa_active_2022}
Lancelot Da~Costa, Samuel Tenka, Dominic Zhao, and Noor Sajid.
\newblock Active {Inference} as a {Model} of {Agency}.
\newblock In \emph{Workshop on {RL} as a model of agency}, 2022{\natexlab{b}}.

\bibitem[Da~Costa et~al.(2024)Da~Costa, Gavenčiak, Hyland, Samiei, Dragos-Manta, Pattisapu, Razi, and Friston]{da_costa_possible_2024}
Lancelot Da~Costa, Tomáš Gavenčiak, David Hyland, Mandana Samiei, Cristian Dragos-Manta, Candice Pattisapu, Adeel Razi, and Karl Friston.
\newblock Possible principles for aligned structure learning agents, September 2024.
\newblock URL \url{http://arxiv.org/abs/2410.00258}.
\newblock arXiv:2410.00258.

\bibitem[Da~Costa et~al.(2025)Da~Costa, Da~Costa, Heins, Medrano, Pavliotis, Parr, Meera, and Friston]{da_costa_theory_2025}
Lancelot Da~Costa, Nathaël Da~Costa, Conor Heins, Johan Medrano, Grigorios~A. Pavliotis, Thomas Parr, Ajith~Anil Meera, and Karl Friston.
\newblock A {Theory} of {Generalized} {Coordinates} for {Stochastic} {Differential} {Equations}.
\newblock \emph{Studies in Applied Mathematics}, 154\penalty0 (5):\penalty0 e70062, 2025.
\newblock ISSN 1467-9590.
\newblock \doi{10.1111/sapm.70062}.
\newblock URL \url{https://onlinelibrary.wiley.com/doi/abs/10.1111/sapm.70062}.
\newblock \_eprint: https://onlinelibrary.wiley.com/doi/pdf/10.1111/sapm.70062.

\bibitem[Deleu et~al.(2023)Deleu, Nishikawa-Toomey, Subramanian, Malkin, Charlin, and Bengio]{deleu2023joint}
Tristan Deleu, Mizu Nishikawa-Toomey, Jithendaraa Subramanian, Nikolay Malkin, Laurent Charlin, and Yoshua Bengio.
\newblock Joint bayesian inference of graphical structure and parameters with a single generative flow network.
\newblock In \emph{Thirty-seventh Conference on Neural Information Processing Systems}, 2023.
\newblock URL \url{https://openreview.net/forum?id=t7lnhhi7De}.

\bibitem[Ding et~al.(2025)Ding, Zhang, Shang, Zhang, Zong, Feng, Yuan, Su, Li, Sukiennik, Xu, and Li]{world_modeling_review}
Jingtao Ding, Yunke Zhang, Yu~Shang, Yuheng Zhang, Zefang Zong, Jie Feng, Yuan Yuan, Hongyuan Su, Nian Li, Nicholas Sukiennik, Fengli Xu, and Yong Li.
\newblock Understanding world or predicting future? a comprehensive survey of world models.
\newblock \emph{ACM Comput. Surv.}, 58\penalty0 (3), September 2025.
\newblock ISSN 0360-0300.
\newblock \doi{10.1145/3746449}.
\newblock URL \url{https://doi.org/10.1145/3746449}.

\bibitem[Ellis et~al.(2015)Ellis, Solar-Lezama, and Tenenbaum]{program_synthesis_1}
Kevin Ellis, Armando Solar-Lezama, and Josh Tenenbaum.
\newblock Unsupervised learning by program synthesis.
\newblock In C.~Cortes, N.~Lawrence, D.~Lee, M.~Sugiyama, and R.~Garnett (eds.), \emph{Advances in Neural Information Processing Systems}, volume~28. Curran Associates, Inc., 2015.
\newblock URL \url{https://proceedings.neurips.cc/paper_files/paper/2015/file/b73dfe25b4b8714c029b37a6ad3006fa-Paper.pdf}.

\bibitem[Ellis et~al.(2019)Ellis, Nye, Pu, Sosa, Tenenbaum, and Solar-Lezama]{program_synthesis_3}
Kevin Ellis, Maxwell Nye, Yewen Pu, Felix Sosa, Joshua~B. Tenenbaum, and Armando Solar-Lezama.
\newblock \emph{Write, execute, assess: program synthesis with a REPL}.
\newblock Curran Associates Inc., Red Hook, NY, USA, 2019.

\bibitem[Fox et~al.(2008)Fox, Sudderth, Jordan, and Willsky]{switching_lds_1}
Emily Fox, Erik Sudderth, Michael Jordan, and Alan Willsky.
\newblock Nonparametric bayesian learning of switching linear dynamical systems.
\newblock In D.~Koller, D.~Schuurmans, Y.~Bengio, and L.~Bottou (eds.), \emph{Advances in Neural Information Processing Systems}, volume~21. Curran Associates, Inc., 2008.
\newblock URL \url{https://proceedings.neurips.cc/paper_files/paper/2008/file/950a4152c2b4aa3ad78bdd6b366cc179-Paper.pdf}.

\bibitem[Friston et~al.(2024{\natexlab{a}})Friston, Heins, Verbelen, Da~Costa, Salvatori, Markovic, Tschantz, Koudahl, Buckley, and Parr]{friston_pixels_2024}
Karl Friston, Conor Heins, Tim Verbelen, Lancelot Da~Costa, Tommaso Salvatori, Dimitrije Markovic, Alexander Tschantz, Magnus Koudahl, Christopher Buckley, and Thomas Parr.
\newblock From pixels to planning: scale-free active inference, July 2024{\natexlab{a}}.
\newblock URL \url{http://arxiv.org/abs/2407.20292}.
\newblock arXiv:2407.20292 [cs, q-bio].

\bibitem[Friston et~al.(2025)Friston, Parr, Heins, Da~Costa, Salvatori, Tschantz, Koudahl, Van~de Maele, Buckley, and Verbelen]{denovo}
Karl Friston, Thomas Parr, Conor Heins, Lancelot Da~Costa, Tommaso Salvatori, Alexander Tschantz, Magnus Koudahl, Toon Van~de Maele, Christopher Buckley, and Tim Verbelen.
\newblock Gradient-free de novo learning.
\newblock \emph{Entropy}, 27\penalty0 (9), 2025.
\newblock ISSN 1099-4300.
\newblock \doi{10.3390/e27090992}.
\newblock URL \url{https://www.mdpi.com/1099-4300/27/9/992}.

\bibitem[Friston et~al.(2024{\natexlab{b}})Friston, Da~Costa, Tschantz, Kiefer, Salvatori, Neacsu, Koudahl, Heins, Sajid, Markovic, Parr, Verbelen, and Buckley]{friston_supervised_2024}
Karl~J. Friston, Lancelot Da~Costa, Alexander Tschantz, Alex Kiefer, Tommaso Salvatori, Victorita Neacsu, Magnus Koudahl, Conor Heins, Noor Sajid, Dimitrije Markovic, Thomas Parr, Tim Verbelen, and Christopher~L. Buckley.
\newblock Supervised structure learning.
\newblock \emph{Biological Psychology}, 193:\penalty0 108891, November 2024{\natexlab{b}}.
\newblock ISSN 0301-0511.
\newblock \doi{10.1016/j.biopsycho.2024.108891}.
\newblock URL \url{https://www.sciencedirect.com/science/article/pii/S0301051124001510}.

\bibitem[Hafner et~al.(2019)Hafner, Lillicrap, Fischer, Villegas, Ha, Lee, and Davidson]{planet}
Danijar Hafner, Timothy Lillicrap, Ian Fischer, Ruben Villegas, David Ha, Honglak Lee, and James Davidson.
\newblock Learning latent dynamics for planning from pixels.
\newblock In \emph{International Conference on Machine Learning}, pp.\  2555--2565, 2019.

\bibitem[Hafner et~al.(2020)Hafner, Lillicrap, Ba, and Norouzi]{dreamer}
Danijar Hafner, Timothy Lillicrap, Jimmy Ba, and Mohammad Norouzi.
\newblock Dream to control: Learning behaviors by latent imagination.
\newblock \emph{ICLR}, 2020.

\bibitem[Hafner et~al.(2025)Hafner, Pasukonis, Ba, and Lillicrap]{dreamerv3}
Danijar Hafner, Jurgis Pasukonis, Jimmy Ba, and Timothy Lillicrap.
\newblock Mastering diverse control tasks through world models.
\newblock \emph{Nature}, pp.\  1--7, 2025.

\bibitem[Hamilton(1990)]{switching_Kalman_1}
James~D. Hamilton.
\newblock Analysis of time series subject to changes in regime.
\newblock \emph{Journal of Econometrics}, 45\penalty0 (1):\penalty0 39--70, 1990.
\newblock ISSN 0304-4076.
\newblock \doi{https://doi.org/10.1016/0304-4076(90)90093-9}.
\newblock URL \url{https://www.sciencedirect.com/science/article/pii/0304407690900939}.

\bibitem[Heins et~al.(2025)Heins, Maele, Tschantz, Linander, Markovic, Salvatori, Pezzato, Catal, Wei, Koudahl, Perin, Friston, Verbelen, and Buckley]{heins_axiom_2025}
Conor Heins, Toon Van~de Maele, Alexander Tschantz, Hampus Linander, Dimitrije Markovic, Tommaso Salvatori, Corrado Pezzato, Ozan Catal, Ran Wei, Magnus Koudahl, Marco Perin, Karl Friston, Tim Verbelen, and Christopher Buckley.
\newblock {AXIOM}: {Learning} to {Play} {Games} in {Minutes} with {Expanding} {Object}-{Centric} {Models}, May 2025.
\newblock URL \url{http://arxiv.org/abs/2505.24784}.
\newblock arXiv:2505.24784 [cs].

\bibitem[Hinton \& Salakhutdinov(2006)Hinton and Salakhutdinov]{gb_rbm}
G.~E. Hinton and R.~R. Salakhutdinov.
\newblock Reducing the dimensionality of data with neural networks.
\newblock \emph{Science}, 313\penalty0 (5786):\penalty0 504--507, 2006.
\newblock \doi{10.1126/science.1127647}.
\newblock URL \url{https://www.science.org/doi/abs/10.1126/science.1127647}.

\bibitem[Hinton et~al.(2006)Hinton, Osindero, and Teh]{dbn}
Geoffrey~E. Hinton, Simon Osindero, and Yee-Whye Teh.
\newblock A fast learning algorithm for deep belief nets.
\newblock \emph{Neural Computation}, 18\penalty0 (7):\penalty0 1527--1554, 07 2006.
\newblock ISSN 0899-7667.
\newblock \doi{10.1162/neco.2006.18.7.1527}.
\newblock URL \url{https://doi.org/10.1162/neco.2006.18.7.1527}.

\bibitem[Isomura et~al.(2023)Isomura, Kotani, Jimbo, and Friston]{isomura}
Takuya Isomura, Kiyoshi Kotani, Yasuhiko Jimbo, and Karl~J. Friston.
\newblock Experimental validation of the free-energy principle with in vitro neural networks.
\newblock \emph{Nature Communications}, 14\penalty0 (4547), 2023.

\bibitem[Jain et~al.(2023)Jain, Deleu, Hartford, Liu, Hernandez-Garcia, and Bengio]{jain_gflownets_2023}
Moksh Jain, Tristan Deleu, Jason Hartford, Cheng-Hao Liu, Alex Hernandez-Garcia, and Yoshua Bengio.
\newblock {GFlowNets} for {AI}-driven scientific discovery.
\newblock \emph{Digital Discovery}, 2\penalty0 (3):\penalty0 557--577, 2023.
\newblock \doi{10.1039/D3DD00002H}.
\newblock URL \url{https://pubs.rsc.org/en/content/articlelanding/2023/dd/d3dd00002h}.
\newblock tex.ids= jainGFlowNetsAIdrivenScientific2023a publisher: Royal Society of Chemistry.

\bibitem[Kalman(1960)]{kalman_filter}
R.~E. Kalman.
\newblock A new approach to linear filtering and prediction problems.
\newblock \emph{Journal of Basic Engineering}, 82\penalty0 (1):\penalty0 35--45, 03 1960.
\newblock ISSN 0021-9223.
\newblock \doi{10.1115/1.3662552}.
\newblock URL \url{https://doi.org/10.1115/1.3662552}.

\bibitem[Kapasi et~al.(2025)Kapasi, Whitehead, and Theogarajan]{gm_rbm}
Nikhil Kapasi, William Whitehead, and Luke Theogarajan.
\newblock The gaussian-multinoulli restricted boltzmann machine: A potts model extension of the grbm, 2025.
\newblock URL \url{https://arxiv.org/abs/2505.11635}.

\bibitem[Koller \& Friedman(2009)Koller and Friedman]{koller_friedman}
Daphne Koller and Nir Friedman.
\newblock \emph{Probabilistic Graphical Models: Principles and Techniques}.
\newblock MIT Press, 2009.

\bibitem[Lanillos et~al.(2021)Lanillos, Meo, Pezzato, Meera, Baioumy, Ohata, Tschantz, Millidge, Wisse, Buckley, and Tani]{lanillos_active_2021}
Pablo Lanillos, Cristian Meo, Corrado Pezzato, Ajith~Anil Meera, Mohamed Baioumy, Wataru Ohata, Alexander Tschantz, Beren Millidge, Martijn Wisse, Christopher~L. Buckley, and Jun Tani.
\newblock Active {Inference} in {Robotics} and {Artificial} {Agents}: {Survey} and {Challenges}.
\newblock \emph{arXiv:2112.01871 [cs]}, December 2021.
\newblock URL \url{http://arxiv.org/abs/2112.01871}.
\newblock arXiv: 2112.01871.

\bibitem[Li et~al.(2025)Li, Key, and Ellis]{program_synthesis_2}
Wen-Ding Li, Darren~Yan Key, and Kevin Ellis.
\newblock Toward trustworthy neural program synthesis.
\newblock In \emph{ICLR 2025 Workshop on Human-AI Coevolution}, 2025.
\newblock URL \url{https://openreview.net/forum?id=HPlvbIJGWy}.

\bibitem[Liao et~al.(2022)Liao, Kornblith, Ren, Fleet, and Hinton]{gb_rbm_2}
Renjie Liao, Simon Kornblith, Mengye Ren, David~J. Fleet, and Geoffrey Hinton.
\newblock Gaussian-bernoulli rbms without tears, 2022.
\newblock URL \url{https://arxiv.org/abs/2210.10318}.

\bibitem[Linderman et~al.(2017)Linderman, Johnson, Miller, Adams, Blei, and Paninski]{linderman_bayesian_2017}
Scott Linderman, Matthew Johnson, Andrew Miller, Ryan Adams, David Blei, and Liam Paninski.
\newblock Bayesian {Learning} and {Inference} in {Recurrent} {Switching} {Linear} {Dynamical} {Systems}.
\newblock In \emph{Proceedings of the 20th {International} {Conference} on {Artificial} {Intelligence} and {Statistics}}, pp.\  914--922. PMLR, April 2017.
\newblock URL \url{https://proceedings.mlr.press/v54/linderman17a.html}.
\newblock ISSN: 2640-3498.

\bibitem[Madsen et~al.(2022)Madsen, Olesen, Jensen, Henriksen, Larsen, and M{\o}ller]{online_conditional_LGS}
Anders~L Madsen, Kristian~G Olesen, Frank Jensen, Per Henriksen, Thomas~M Larsen, and J{\o}rn~M M{\o}ller.
\newblock Online updating of conditional linear gaussian bayesian networks.
\newblock In Antonio Salmeron and Rafael Rumi (eds.), \emph{Proceedings of The 11th International Conference on Probabilistic Graphical Models}, volume 186 of \emph{Proceedings of Machine Learning Research}, pp.\  97--108. PMLR, 05--07 Oct 2022.
\newblock URL \url{https://proceedings.mlr.press/v186/madsen22a.html}.

\bibitem[Murphy(2001)]{murphy_active_nodate}
Kevin Murphy.
\newblock Active {Learning} of {Causal} {Bayes} {Net} {Structure}.
\newblock 2001.

\bibitem[Murphy(1998)]{murphy_1998_switching}
Kevin~P. Murphy.
\newblock Switching kalman filters.
\newblock Technical report, University of British Columbia, Vancouver, Canada, 1998.
\newblock Available online at \url{https://www.cs.ubc.ca/~murphyk/Papers/skf.pdf}.

\bibitem[Parr \& Friston(2018)Parr and Friston]{parr_active_2018}
Thomas Parr and Karl~J. Friston.
\newblock Active inference and the anatomy of oculomotion.
\newblock \emph{Neuropsychologia}, 111:\penalty0 334--343, March 2018.
\newblock ISSN 0028-3932.
\newblock \doi{10.1016/j.neuropsychologia.2018.01.041}.
\newblock URL \url{http://www.sciencedirect.com/science/article/pii/S0028393218300472}.

\bibitem[Parr et~al.(2021)Parr, Limanowski, Rawji, and Friston]{parr_computational_2021}
Thomas Parr, Jakub Limanowski, Vishal Rawji, and Karl Friston.
\newblock The computational neurology of movement under active inference.
\newblock \emph{Brain}, 144\penalty0 (6):\penalty0 1799--1818, June 2021.
\newblock ISSN 0006-8950.
\newblock \doi{10.1093/brain/awab085}.
\newblock URL \url{https://doi.org/10.1093/brain/awab085}.
\newblock tex.ids= parrComputationalNeurologyMovement2021.

\bibitem[Parr et~al.(2022)Parr, Pezzulo, and Friston]{parr_active_2022}
Thomas Parr, Giovanni Pezzulo, and Karl~J. Friston.
\newblock \emph{Active {Inference}: {The} {Free} {Energy} {Principle} in {Mind}, {Brain}, and {Behavior}}.
\newblock MIT Press, Cambridge, MA, USA, March 2022.
\newblock ISBN 978-0-262-04535-3.

\bibitem[Pezzato et~al.(2025)Pezzato, Çatal, Maele, Pitliya, and Verbelen]{pezzato_mobile_2025}
Corrado Pezzato, Ozan Çatal, Toon Van~de Maele, Riddhi~J. Pitliya, and Tim Verbelen.
\newblock Mobile {Manipulation} with {Active} {Inference} for {Long}-{Horizon} {Rearrangement} {Tasks}, July 2025.
\newblock URL \url{http://arxiv.org/abs/2507.17338}.
\newblock arXiv:2507.17338 [cs].

\bibitem[Pu et~al.(2020)Pu, Ellis, Kryven, Tenenbaum, and Solar-Lezama]{program_synthesis_4}
Yewen Pu, Kevin Ellis, Marta Kryven, Josh Tenenbaum, and Armando Solar-Lezama.
\newblock Program synthesis with pragmatic communication, 2020.
\newblock URL \url{https://arxiv.org/abs/2007.05060}.

\bibitem[Rabiner \& Juang(1986)Rabiner and Juang]{rabiner_hmm_intro}
L.~Rabiner and B.~Juang.
\newblock An introduction to hidden markov models.
\newblock \emph{IEEE ASSP Magazine}, 3\penalty0 (1):\penalty0 4--16, 1986.
\newblock \doi{10.1109/MASSP.1986.1165342}.

\bibitem[Rabiner(1989)]{rabiner_hmm_applications}
L.R. Rabiner.
\newblock A tutorial on hidden markov models and selected applications in speech recognition.
\newblock \emph{Proceedings of the IEEE}, 77\penalty0 (2):\penalty0 257--286, 1989.
\newblock \doi{10.1109/5.18626}.

\bibitem[Schrittwieser et~al.(2020)Schrittwieser, Antonoglou, Hubert, Simonyan, Sifre, Schmitt, Guez, Lockhart, Hassabis, Graepel, Lillicrap, and Silver]{muzero}
Julian Schrittwieser, Ioannis Antonoglou, Thomas Hubert, Karen Simonyan, Laurent Sifre, Simon Schmitt, Arthur Guez, Edward Lockhart, Demis Hassabis, Thore Graepel, Timothy Lillicrap, and David Silver.
\newblock Mastering {Atari}, {Go}, {Chess} and {Shogi} by {Planning} with a {Learned} {Model}.
\newblock \emph{Nature}, 588\penalty0 (7839):\penalty0 604–609, December 2020.
\newblock ISSN 1476-4687.
\newblock \doi{10.1038/s41586-020-03051-4}.
\newblock URL \url{http://dx.doi.org/10.1038/s41586-020-03051-4}.

\bibitem[Schwarzer et~al.(2023)Schwarzer, Obando-Ceron, Courville, Bellemare, Agarwal, and Castro]{bbf}
Max Schwarzer, Johan Obando-Ceron, Aaron Courville, Marc~G. Bellemare, Rishabh Agarwal, and Pablo~Samuel Castro.
\newblock Bigger, better, faster: human-level atari with human-level efficiency.
\newblock In \emph{Proceedings of the 40th International Conference on Machine Learning}, ICML'23. JMLR.org, 2023.

\bibitem[Winn \& Bishop(2005)Winn and Bishop]{winn_variational_2005}
John Winn and Christopher~M Bishop.
\newblock Variational {Message} {Passing}.
\newblock \emph{Journal of Machine Learning Research}, pp.\ ~34, 2005.

\bibitem[Yu(2010)]{hidden_semi_markov}
Shun-Zheng Yu.
\newblock Hidden semi-markov models.
\newblock \emph{Artificial Intelligence}, 174\penalty0 (2):\penalty0 215--243, 2010.
\newblock ISSN 0004-3702.
\newblock \doi{https://doi.org/10.1016/j.artint.2009.11.011}.
\newblock URL \url{https://www.sciencedirect.com/science/article/pii/S0004370209001416}.
\newblock Special Review Issue.

\bibitem[Yu et~al.(2019)Yu, Chen, Gao, and Yu]{dag_gnn}
Yue Yu, Jie Chen, Tian Gao, and Mo~Yu.
\newblock Dag-gnn: Dag structure learning with graph neural networks.
\newblock In \emph{Proceedings of the 36th International Conference on Machine Learning}, 2019.

\bibitem[Zheng et~al.(2018)Zheng, Aragam, Ravikumar, and Xing]{notears}
Xun Zheng, Bryon Aragam, Pradeep Ravikumar, and Eric~P. Xing.
\newblock Dags with no tears: continuous optimization for structure learning.
\newblock In \emph{Proceedings of the 32nd International Conference on Neural Information Processing Systems}, NIPS'18, pp.\  9492–9503, Red Hook, NY, USA, 2018.

\bibitem[Çatal et~al.(2021)Çatal, Verbelen, Van~de Maele, Dhoedt, and Safron]{catal_robot_2021}
Ozan Çatal, Tim Verbelen, Toon Van~de Maele, Bart Dhoedt, and Adam Safron.
\newblock Robot navigation as hierarchical active inference.
\newblock \emph{Neural Networks}, 142:\penalty0 192--204, October 2021.
\newblock ISSN 0893-6080.
\newblock \doi{10.1016/j.neunet.2021.05.010}.
\newblock URL \url{https://www.sciencedirect.com/science/article/pii/S0893608021002021}.

\end{thebibliography}

\appendix

\section{Related Work}
\label{app:related}

\noindent\textbf{Neural World Models.} Recent advances in model-based reinforcement learning have produced impressive neural world models.
Dreamer \citep{dreamer} learns latent dynamics models using recurrent state-space models, while PlaNet \citep{planet} pioneered planning with learned models directly from pixels.
MuZero \citep{muzero} learns models implicitly through value-equivalent predictions.
While these achieve strong performance, they operate as black boxes, making them difficult to interpret, debug, or trust in safety-critical applications. 
Our approach emphasizes interpretability and uncertainty quantification while maintaining competitive performance through the usage of sparse hierarchies of building blocks that support Bayesian inference \citep{friston_pixels_2024}. Additionally, the incremental structure learning by starting from a minimal model and then growing (FSL) introduces a bias toward minimal models that are more parameter efficient. In particular, the hierarchical sLDS developed in AXIOM demonstrates comparable performance to standard model-based reinforcement learning approaches while being far more parameter and sample efficient \citep{heins_axiom_2025}.

\noindent\textbf{Structured Probabilistic Models.} Classical work on structured models includes Hidden Markov Models \citep{rabiner_hmm_intro, rabiner_hmm_applications}, Kalman filters \citep{kalman_filter}, deep belief networks \citep{dbn}, and their switching variants \citep{linderman_bayesian_2017}.
These models offer interpretability but traditionally struggled with the complexity of real-world domains.
Our framework extends these classical models through hierarchical composition and four types of structural depth, achieving higher expressiveness. The extent to which this matches what is needed for large scale applications remains to be investigated.

\noindent\textbf{Hybrid Discrete-Continuous Models.} The combination of discrete and continuous dynamics has been explored in various contexts including hidden semi-Markov models \citep{hidden_semi_markov}, conditional linear Gaussian models \citep{online_conditional_LGS}, switching Kalman filters \citep{murphy_1998_switching, switching_Kalman_1}, mixture density networks \citep{mixture_density}, Gaussian-Bernoulli restricted Boltzmann machines \citep{gb_rbm} and its extensions \citep{gm_rbm, gb_rbm_2}, and the classic sLDS model \citep{switching_lds_2}.
Recurrent switching linear dynamical systems (rsLDS) \citep{linderman_bayesian_2017} represent a key advance, allowing continuous states to influence discrete mode switching.
Motivated by arguments from stochastic process theory, we propose to systematically combine these with hierarchical POMDPs to create a principled, highly expressive framework that combines the complementary strengths of discrete and continuous state-spaces. 

\noindent\textbf{Structure Learning.} Traditional structure learning in graphical models faces combinatorial explosion \citep{koller_friedman}.
Our framework focuses on time-series data, and uses this inductive bias to meaningfully limit the graphical structure, hence finessing the combinatorial explosion while maintaining expressiveness. Scalable joint structure and parameter learning in this model class is still an open challenge, and we anticipate recent advances in structure learning 
such as continuous relaxations and differentiable structure learning \citep{notears, dag_gnn} or generative flow networks \citep{deleu2023joint, joint_structure_params} to be useful here. The FSL algorithm \citep{friston_pixels_2024} offers a novel solution to this problem by growing structure and parameters incrementally together, which is currently limited in scale.

\noindent\textbf{Program Synthesis and Induction.} Work on program synthesis \citep{program_synthesis_1, program_synthesis_2, program_synthesis_3, program_synthesis_4} searches over general program spaces to explain data.
While powerful in principle, the space of general programs is vast and often intractable.
Our approach differs by constraining search to natural stochastic primitives (HMMs, sLDS) that are fundamental to world modeling, making the search space more tractable while maintaining sufficient expressiveness.

\noindent\textbf{Active Inference.} Our approach follows a line of research on active inference \citep{da_costa_active_2020,parr_active_2022,friston_supervised_2024}. Active inference agents usually employ hierarchies of continuous and discrete POMDP world models with variational Bayesian inference to infer states and parameters. Thus, the active inference approach aligns with our natural building blocks.

\section{Detailed Building Block Descriptions}
\label{app:math}

\subsection{Hidden Markov Models (HMMs) and POMDPs}

Hidden Markov Models capture categorical dynamics through finite sets of latent states $s \in \mathcal{S}$ and observations $o \in \mathcal{O}$. The base model specifies transition probabilities $P(s_{t+1}|s_t)$ and observation probabilities $P(o_t|s_t)$, providing a fundamental model for time series. When augmented with extra latent states (i.e., generalized depth, see Section \ref{sec:depth}) that are controllable---denoted $a \in \mathcal{A}$---the HMM transforms into a POMDP with action-dependent transitions $P(s_{t+1}|s_t, a_t)$. The result is a model that encompasses a wide class of problem types involving the control of latent states in a dynamical system represented in terms of higher-order categorical variables. This enables perception \emph{and} action; the same discrete-state machinery that supports passive inference over latent causes also supports active planning and environmental control for agent modeling.

HMMs naturally express all Markov processes in discrete latent spaces: Markov chains. 
When augmented with generalized depth, they can express all latent stochastic processes with finite memory. Crucially, \emph{discrete models are more universal than continuous ones}---any continuous space can be discretized through quantization.
By discretizing continuous spaces, HMMs can approximate a wide class of continuous stochastic processes in latent space, not limited to SDEs. 
However, this universality comes at a cost: they scale poorly to high-dimensional spaces. Consequently, effective applications rely on structured state spaces (factorization, hierarchy, sparsity) and problem-specific abstractions to keep inference and learning tractable.

\subsection{Recurrent Switching Linear Dynamical Systems (rsLDS)}

The sLDS class of models provides a continuous counterpart to HMMs for time series modeling, approximating nonlinear dynamics through switching mixtures of linear ones.
A key innovation is the \emph{recurrent} connection \citep{linderman_bayesian_2017}: continuous states influence discrete switching, essential for modeling events like bouncing balls where contact triggers a switching of dynamics.
The model maintains discrete switch states $s_t^{d} \in \{1, ..., K\}$ and continuous states $s_t^{c} \in \mathbb{R}^D$.
For passive modeling dynamics unfold as follows: 
\[
s_{t+1}^{\mathrm{c}} = B_{s_{t+1}^{\mathrm{d}}}\, s_t^{\mathrm{c}} + \epsilon_t, 
\quad \epsilon_t \sim \mathcal{N}(0, \Sigma_{z_t}).
\] 

While adding control inputs $a_t$ through generalized depth enables active modeling:
\[
s_{t+1}^{c} = B_{s_{t+1}^{d}}^{s}s_{t}^{c} +B_{a_{t+1}^{d}}^{a}a_{t}^{c} + \epsilon_t.
\]

Recurrent sLDS approximate diffusion processes (SDEs) through Euler discretization and switching linear dynamics.
While more restricted than discrete models (limited to SDEs), they scale more efficiently to high dimensions, making them the natural choice for physical modeling and low-level control. Generalized coordinates (currently not implemented for rsLDS) endows the process with memory, enabling approximation of non-Markovian SDEs such as encountered in complex systems modeling.

\subsection{Hierarchical Composition}

Hierarchical composition captures phenomena at different spatial and temporal scales, with hierarchies extending the expressiveness properties of the component layers to multi-scale phenomena. 
Higher layer states provide initial conditions for lower layer trajectories, enabling multi-scale modeling; e.g. from abstract planning to detailed execution.
The framework currently supports three composition patterns: discrete→discrete (logical hierarchies), discrete→continuous (plans to physics), and continuous→continuous (multi-scale physics).
Note that continuous→discrete composition may not be practically useful, as high-level discrete reasoning naturally contextualizes concrete low-level dynamics.

\subsection{Tractable Structure Learning} \label{subsec:inf_learning_detail}

Traditional structure learning faces a fundamental computational barrier: the number of possible directed acyclic graph models (DAGs) grows super-exponentially with the number of variables.
The number of DAGs with $n$ unlabeled nodes grows combinatorially: For just $n=10$ variables, there are already $>10^{12}$ possible DAGs, making exhaustive search extremely challenging. Our approach sidesteps this explosion by fixing the causal structure (hierarchical HMMs/sLDS) and searching only over depth parameters (hierarchical, and per layer temporal, factorial and generalized depth).
This reduces the search space dramatically while maintaining expressiveness---the causal structure is largely determined by the choice of representations.

Even with fixed compositional structure, learning parameters in deep models remains challenging.
Full Bayesian learning of transition and likelihood parameters in deep POMDPs without prior knowledge is difficult.
FSL offers a novel solution by growing structure and parameters \emph{jointly and incrementally}: crucially, it learns parameters for each newly added component while simultaneously expanding the structure, finessing the difficulty of parameter learning in deep models that would otherwise be very challenging. 
While very efficient, the current implementation of FSL \cite{friston_pixels_2024} is limited in scale since it yields generative models that are a quasi-\emph{lossless compression} of the data. Furthermore, it is so far limiting to learning the structure of models under scale-invariant inductive biases, which may be a limitation, although this is sufficient to model phenomena that exhibit self-similarity in structure across different spatiotemporal scales, which perhaps surprisingly may include video and audio streams.
Looking forward, growing structure and parameters \emph{jointly and incrementally} seems promising for learning deep Bayesian models.

\section{Empirical Evidence Descriptions}
\label{app:empirical}

\subsection{Multimodal Generative Modeling}
\label{app:passive}
Figure \ref{fig:multimodal} demonstrates multimodal generation using only our structured building blocks, without any neural networks. The underlying learned model, a renormalizing generative model (RGM), is a hierarchical HMM with scale-invariant structural constraints \citep{friston_pixels_2024}---no actions are needed for this passive modeling task. While current scale is limited by structure learning algorithms, these results demonstrate that pure probabilistic inference with interpretable components can achieve multimodal expressiveness. The video generation captures realistic bird motion patterns, while the audio generation produces structured jazz-like musical sequences, both emerging from the same underlying hierarchical HMM framework with different observation modalities. When actions are added to the RGM it can be used to learn the necessary paths (sequences of actions that transition among states) to recapitulate expert play in object-centric arcade style games \citep{friston_pixels_2024, denovo}.

RGMs are based on the prior assumption that the data has the same self-similarity structure at each levels of abstraction. The result is a hierarchical HMM model (or POMDP in the case where actions are present) that assumes spatiotemporal dynamics are the same across different levels of the model but viewed at different scales with different parameters and observations at each level. This self-sameness property which is present at successive layers is found in many natural phenomena, as is found in the renormalization group (RG), whose operator is responsible for performing coarse-graining and scaling operations used in the hierarchical layers of the model.

Note that the original paper \cite{friston_pixels_2024} assumes that expert play is available (the ``Maxwell's daemon" approach) but in \citep{denovo} the model can be built from random sequences of observations.

The renormalization operation is performed in both space and time. The temporal renormalization results in a model where states at one level constrain the initial latent states and paths (random variables in the model which are sequences of actions that transition among latent states) at the level below thus controlling their temporal dynamics. This can be thought of as a version of a discrete switching dynamical system where paths function as switching variables that change at a slower timescale than the paths at the hierarchical level being switched. The result is a discrete semi-Markovian structure of sequences of sequences. Spatial renormalization results because groups of states at one level of the hierarchy are generated by a single state at the level above. This results in a structure where latent states are conditionally independent at every level and which generate local dependencies within each group of states at the lower level.

The FSL algorithm is used to build the RGM. First, the FSL algorithm quantizes continuous time-color pixel values for an image or frame into a discrete state-space. In the original RGM paper \citep{friston_pixels_2024}, the image is spatially partitioned into overlapping patches (tiles) where each pixel is assigned a weight based on a Gaussian radial function that specifies its distance from a patch centroid to capture local interactions between pixels. Each sequence of patches is then subject to a singular value decomposition (SVD) to provide a compressed representation of how the spatial content of each patch changes over time (captured in the left singular vectors). Spatial renormalization then proceeds by recursively grouping patches in each frame into neighboring quantized spatial blocks such that the image resolution is halved (coarse-grained). 

At the lowest level of the model, a blocking transformation is used to group pixels together. The next level is built by defining a latent state as a unique pattern of pixel observations that are frequently observed together, suggesting that they are generated by the same initial state and path. Neighboring states and paths are then grouped together and the resulting block is defined to control the initial conditions of the states at the level below. In \citep{denovo} an alternative grouping operation is used based on the pairwise mutual information between time-varying pixels. This is based on the notion of local interactions which implicitly assume that pixels that change in the same way ought to be grouped together. The likelihoods in the model (combinations of parameters specifying the distributions over initial states and paths) accumulate Dirichlet counts until the mutual information between states and observations at the hierachical level converges to its maximum. The blocking procedure is recursively applied until only one group remains at the highest level at which point the FSL algorithm terminates and the RGM has been fully constructed.

\subsection{Planning from Pixels}
\label{app:active}
\citet{heins_axiom_2025} introduced AXIOM  (\textbf{A}ctive e\textbf{X}panding \textbf{I}nference with \textbf{O}bject-centric \textbf{M}odels), which demonstrated better planning performance in simplified Atari-style games using hierarchical continuous rsLDS with control inputs.

The model achieves competitive performance with state-of-the-art neural methods (Dreamer V3, BBF) while maintaining interpretability.
Notably, consistently better early learning curves demonstrate superior sample efficiency.

AXIOM relies upon the notion of object-centric core priors which model the world as discrete objects with sparse interactions and locally (piecewise) linear dynamics of the kind found in arcade game environments. The AXIOM model is a POMDP where latent states, $\mathcal{Z}_t$, change due to actions taken by the agent at a particular time point. Using image frames as observations, $y_t$, and a reward $r_t$, AXIOM learns a state-space model with object-centric priors via variational inference. The full generative model for a sequence of states, observations, and parameters ($\boldsymbol{\Theta}$), is represented as follows:

\begin{equation}
    \begin{split}
        p(\boldsymbol{y}_{0:T}, \mathcal{Z}_{0:T}, \tilde{\boldsymbol{\Theta}}) = p(y_0, \mathcal{Z}_0)p(\tilde{\boldsymbol{\Theta}}) &\prod_{t=1}^T 
        \underbrace{p(\boldsymbol{x}_{t-1} \mid \boldsymbol{z}_t, \boldsymbol{\Theta}_{\text{iMM}})}_{\text{Identity mixture model}}  \underbrace{p(\boldsymbol{x}_{t-1}, \boldsymbol{z}_t, \boldsymbol{s}_t, a_{t-1}, r_t \mid \boldsymbol{\Theta}_{\text{rMM}})}_{\text{Recurrent mixture model}} \\ &\prod_{k=1}^K 
        \underbrace{p(\boldsymbol{y}_t \mid \boldsymbol{x}_t^{(k)}, \boldsymbol{z}_{t, \text{sMM}}, \boldsymbol{\Theta}_{\text{sMM}})}_{\text{Slot mixture model}} \underbrace{p(\boldsymbol{x}_t^{(k)} \mid \boldsymbol{x}_{t-1}^{(k)}, \boldsymbol{s}_t^{(k)}, \boldsymbol{\Theta}_{\text{tMM}})}_{\text{Transition mixture model}}
    \end{split}
\end{equation}

where discrete latent states are represented as two subtypes: $\boldsymbol{z}_t^{(k)}$ and $\boldsymbol{s}_t^{(k)}$ which capture object features and pairs of switch states (corresponding to the object feature's trajectory) respectively. As the probabilistic model implies, there are four modules with distinct roles:

\begin{enumerate}
    \item \textbf{Slot mixture model (sMM)}: The sMM explains pixels with different slots. These slots correspond to object features such shape, color, or position that compete to decide which feature best explains a pixel in the input frame. The slot mixture model can grow dynamically to add or remove slots with each new observation until all relevant features have been learned.
    \item \textbf{Identity mixture model (iMM)}: The iMM is responsible for inferring an identity code or type for each object in image frame based on its features so that the objects' dynamics can be inferred.
    \item \textbf{Transition mixture model (tMM)}: The tMM is an sLDS that captures the transition probabilities among object states which corresponds to object motion.
    \item \textbf{Recurrent mixture model (rMM)}: The rMM infers the switch states of the tMM from the object slots in order to model the interaction between objects based on the most recent action and rewards. This can be used to predict the next switching state of the tMM.
\end{enumerate}

All modules can adaptively grow or shrink online as data arrives based on the log evidence for the new observation.

\section{Future Directions}
\label{app:future}

\subsection{Long-Term Vision}

We have presented a framework for building expressive world models that can handle a large class of problems. The framework consists of natural building blocks that we believe could become a foundational infrastructure for world modeling. Our empirical data demonstrates the flexibility of the models in a variety of domains including hybrid models and multimodal data streams. Our framework maintains higher interpretability than neural network-based models. The combination of interpretability, uncertainty quantification, and expressiveness at scale would unlock applications in safety-critical domains, scientific discovery, and open-ended learning.

We envision this framework as providing the same sort of modularity Keras and other deep learning frameworks provided in deep learning, enabling the community to build upon a common foundation, standardizing world model components to accelerate progress across the field. We also envision our framework as having great utility in the space of scientific discovery which is inherently hypothesis-driven and does not directly rely on rewards. However, this application has not yet been analyzed in detail. Despite the promise of our framework, some open questions and challenges still remain.

\subsection{Open questions and challenges}

\noindent\textbf{Structure Learning and scaling limitations.} The joint inference of structures and parameters is still an open challenge.

One promising avenue casts structure and parameter inference as hierarchical Bayesian inference, possibly leveraging uncertainty in action selection to actively learn these with minimal
  samples \citep{murphy_active_nodate,da_costa_possible_2024}. 
This requires
  solving deep Bayesian parameter learning in
  these hierarchies—currently an outstanding
  challenge although there are many promising directions, e.g. utilizing neural network approaches \citep{deleu2023joint, joint_structure_params}.

As mentioned previously, the FSL algorithm \citep{friston_pixels_2024} is a fast and interpretable approach to solving the problem of joint inference over structures and parameters. However, FSL has a number of limitations that must be overcome for it to be usefully applied at scale. (1) The current implementation has not yet been stress-tested for scalability, yet the method itself is currently bottlenecked by implementing a quasi-lossless compression of the data, which severely limits its applicability in real-world applications. More scalable alternatives need to be investigated. (2) FSL is limited to discrete state-space HMM/POMDP models and imposes a renormalization constraint into the structure of the model which may not be suitable in all circumstances. (3) FSL does not provide uncertainty over structures themselves which precludes active learning via structural information gain.

\noindent\textbf{Architectural limitations.} The sLDS model, one of the natural building blocks in our framework, naturally expresses partially-observed Markovian SDE models. Yet, current implementations do not yet support generalized depth which would enable expressing partially-observed semi-Markovian SDEs. Extensions into this domain will address some theoretical expressiveness gaps for the sLDS model.

We have provided empirical examples of the wide applicability of the natural building blocks in our framework. However, to date there is no published literature testing the expressivity of architectures combining the building blocks along all depth parameters.

\noindent\textbf{Theoretical and applied open questions.} Our framework provides a step toward a universally expressive model space suitable for open-ended learning. Future work will identify its theoretical limits in expressiveness and identify any expressiveness gaps, which will in turn motivate future development. 
Finally, we hypothesize that because our framework is more highly interpretable than black box approaches and allows for uncertainty quantification, it is well suited for applications in mission and safety-critical domains. Future work will be needed to provide empirical evidence to more critically evaluate the validity of these claims.

\end{document}